# A LongFormer-Based Framework for Accurate and Efficient Medical Text Summarization

Dan Sun
Washington University in St. Louis
St. Louis, USA

Jacky He
Cornell University
New York, USA

Hanlu Zhang
Stevens Institute of Technology
Hoboken, USA

Zhen Qi
Northeastern University
Boston, USA

Hongye Zheng
The Chinese University of Hong Kong
Hong Kong, China

Xiaokai Wang *
Santa Clara University
Santa Clara, USA

*Abstract-This paper proposes a medical text summarization method based on LongFormer, aimed at addressing the challenges faced by existing models when processing long medical texts. Traditional summarization methods are often limited by short-term memory, leading to information loss or reduced summary quality in long texts. LongFormer, by introducing long-range self-attention, effectively captures long-range dependencies in the text, retaining more key information and improving the accuracy and information retention of summaries. Experimental results show that the LongFormer-based model outperforms traditional models, such as RNN, T5, and BERT in automatic evaluation metrics like ROUGE. It also receives high scores in expert evaluations, particularly excelling in information retention and grammatical accuracy. However, there is still room for improvement in terms of conciseness and readability. Some experts noted that the generated summaries contain redundant information, which affects conciseness. Future research will focus on further optimizing the model structure to enhance conciseness and fluency, achieving more efficient medical text summarization. As medical data continues to grow, automated summarization technology will play an increasingly important role in fields such as medical research, clinical decision support, and knowledge management.*

I. INTRODUCTION

With the continuous development of the medical field, the volume of scientific literature is increasing rapidly. Medical professionals are facing growing pressure to process this vast amount of information. Medical literature not only covers a wide range of domain knowledge but also includes experimental data, case studies, and treatment methods [1]. Traditional literature retrieval and reading methods can no longer meet the needs of doctors and researchers for efficient access to useful information [2]. Therefore, how to efficiently extract core content from the massive volume of medical literature and automatically generate concise and accurate summaries has become an important issue in current medical information processing.

In the medical field, summary generation is a crucial task in natural language processing (NLP). The goal of automatic summarization is to extract the most informative parts from a long medical article and generate a concise, easy-to-understand summary while retaining key information. Traditional summarization methods include rule-based approaches, statistical methods, and early machine-learning techniques. However, these methods often rely on manual feature extraction or simple sentence-extraction strategies [3]. They generally fail to understand the deeper semantics of the text and cannot handle complex contextual information in long documents. With the rapid development of deep learning technologies, deep learning-based automatic summarization methods have gradually become the mainstream research direction. These methods, through pretraining on large-scale corpora, capture more complex linguistic features and improve the quality and accuracy of summarization [4].

However, despite significant progress in summarization through deep learning-based models, there are still challenges, especially when handling long texts. Many existing pre-trained language models (such as BERT and GPT) face difficulties in capturing long-range dependencies in long texts due to input sequence length limitations. Medical literature is typically lengthy and contains complex terminology and context, which makes it difficult for existing models to process effectively. Therefore, researching how to handle long medical texts and generate high-quality summaries becomes an urgent issue.

To address this challenge, the LongFormer model, a novel approach for handling long texts, has shown distinct advantages. LongFormer introduces a sparse attention mechanism, allowing the model to efficiently process longer input sequences than traditional Transformer models [5]. Unlike traditional global attention mechanisms, LongFormer combines local windows and global sparse attention, effectively reducing computational complexity while maintaining performance [6], which has been widely leveraged in the computer vision [7] and medical diagnosis fields [8]. This allows LongFormer to handle long texts and capture long-range dependencies without sacrificing efficiency. This characteristic makes LongFormer an ideal choice for generating medical literature summaries. It can efficiently process medical texts while fully understanding contextual information, and generating highly accurate summaries [9].

Through this research, we aim to provide a new solution for medical literature summarization and drive the further development of deep learning technologies in the medical field. Specifically, in the face of the challenge posed by the vast amount of medical literature, the LongFormer-based summarization method can provide doctors, researchers, and other professionals with an efficient and accurate tool to save time and improve decision-making efficiency. With the continuous advancement of AI technologies, the medical field will eventually realize smarter information processing and decision support, leading to a more efficient and precise healthcare system.

## II. RELATED WORK

### A. LongFormer

LongFormer is a Transformer-based model designed to address the computational bottleneck that traditional Transformers face when processing long texts [10]. Traditional Transformer models use global self-attention to process each word in the input. However, their computational complexity grows quadratically with the input length, making them highly inefficient for long-text processing. To resolve this issue, LongFormer introduces a sparse attention mechanism that combines a sliding window (local window) with global attention. This allows each word's attention to be calculated based only on its local neighborhood and a few global tokens, significantly reducing the computational load. This sparse attention mechanism not only improves efficiency in processing long texts but also preserves the model's ability to capture long-range dependencies, enabling LongFormer to perform exceptionally well on long documents [11].

Since its introduction, LongFormer has been widely applied to various tasks, including long-text classification, question-answering, and document summarization. Compared to traditional models, LongFormer has shown significant advantages on several benchmark datasets, especially in processing lengthy texts. Due to its efficient computational structure, LongFormer can effectively extract key information from complex, lengthy texts, such as medical literature and legal documents. Research based on this model has gradually gained attention in academia, and it is considered a powerful tool for solving the long-text processing problem. The advantages of LongFormer make it particularly suitable for tasks that require modeling large amounts of contextual information, providing new possibilities for the application of deep learning models across various fields [12].

### B. Medical text summary generation

Medical text summarization is an important research direction in the field of natural language processing, especially when dealing with vast amounts of medical literature and electronic medical records. Efficiently and accurately extracting valuable information has become a focal point for both academia and industry. Early studies were mainly based on traditional statistical methods and rule-based extractive summarization techniques, such as TF-IDF and TextRank. These methods construct summaries by selecting keywords or sentences with high weights. While effective in some specific scenarios, they often overlook the complex semantics and long-range dependencies present in medical texts [13].

In recent years, Transformer-based models, such as BERT, GPT, and BART, have been widely applied to medical text summarization tasks due to their superior ability to model context. These models, through pretraining and fine-tuning, capture complex terminology and syntactic structures in medical texts, improving both the accuracy and fluency of the generated summaries. For example, BERT and its variants like BioBERT and ClinicalBERT have achieved significant results in various medical tasks, while BART, due to its powerful sequence-to-sequence generation capabilities, has excelled in automatic medical literature summarization.

Furthermore, with the increasing diversity of medical data, new approaches are gradually incorporating multimodal data (such as combining text with medical images) into summarization tasks to further enhance the comprehensiveness and precision of the summaries. However, despite these advancements, current models still face computational bottlenecks when processing extremely long medical documents. Therefore, improving computational efficiency while maintaining summarization quality remains a major challenge in current research.

## III. METHOD

In order to effectively generate medical text summaries, this study proposed a LongFormer-based model to process long medical documents through a sparse attention mechanism and combined it with a sequence-to-sequence generation method to achieve summary generation [14].

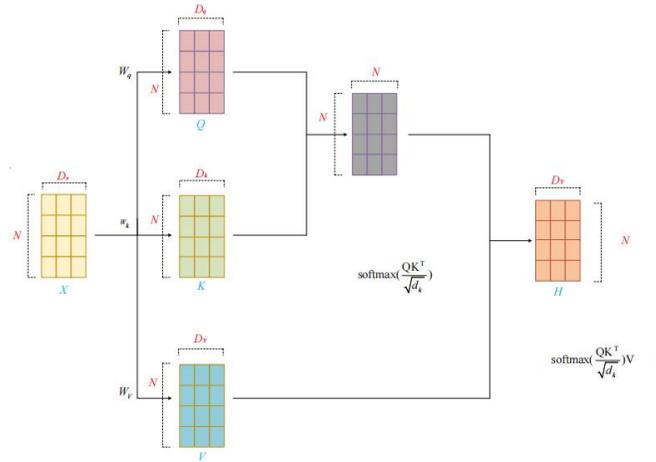

Figure 1 LongFormer overall network architecture

Figure 1 shows the overall architecture of the LongFormer model, which uses a sparse attention mechanism to process long text data. Through linear transformation, the input data is converted into queries, keys, and values, and the attention output is generated by calculating the relationship between them. Finally, the model uses the features generated by these attention mechanisms for summary generation, optimizing the efficiency of processing long documents.

During the entire experiment, first, for the input long text $X = \{x_1, x_2, ..., x_n\}$ where $x_i$ represents the i-th in the text, the model's goal is to generate a concise summary $Y = \{y_1, y_2, ..., y_m\}$, where m is the length of the summary. In order to process long texts, LongFormer uses a sparse attention mechanism to replace the global attention mechanism in the traditional Transformer [15]. For the attention calculation of each layer, LongFormer calculates the attention weight at each position i as follows:

$$A(i,j) = \frac{\exp(\frac{Q_i K_j^T}{\sqrt{d_k}})}{\sum_{k \in N_i} \exp(\frac{Q K_k^T}{\sqrt{d_k}})}$$

Where $Q_i$ and $K_i$ represent the query and key vectors at the i-th and j-th positions in the input sequence, respectively. $d_k$ is the dimension of the key vector. $N_i$ is the attention range at position i, which usually includes a fixed local window and some global tags[16]. By introducing this sparse attention mechanism, LongFormer can significantly reduce the computational complexity and thus process longer input sequences. Its sparse attention mechanism is shown in Figure 2

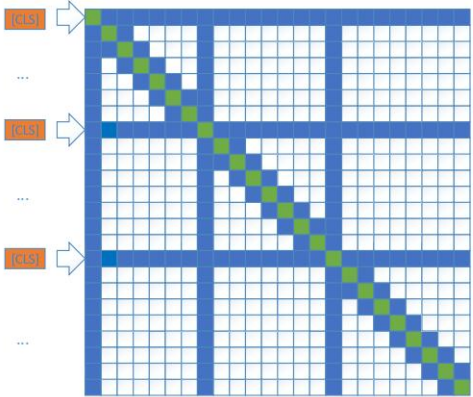

Figure 2 Sparse Attention Mechanism

Next, after passing through multiple Transformer encoding layers, we get the context representation $H = \{h_1, h_2, ..., h_n\}$ of each word. In order to generate summaries from these context representations, we use a decoder that adopts a sequence-to-sequence generation framework similar to BART [17]. The decoder generates the corresponding summary vocabulary based on the output of the encoder. During the decoding process, given the probability distribution of the previous vocabulary, the generation probability of the current vocabulary can be expressed as:

$$P(y_t | y_{<t}, X) = \text{softmax}(W_h h_t + b)$$

Among them, $y_t$ is the generated word of the decoder at the current time step, $h_t$ is the context representation output by the encoder, $W_h$ and b are the weight matrix and bias term respectively. By maximizing this probability distribution, we can gradually generate each word in the summary.

During training, we optimize the model by minimizing the loss between the predicted summary and the true summary. A commonly used loss function is the cross-entropy loss, which is defined as:

$$L = -\sum_{t=1}^{m} \log P(y_t | y_{<t}, X)$$

Where $y_t$ is the t-th word in the real summary, and $P(y_t | y_{<t}, X)$ is the probability of the word predicted by the model. By minimizing this loss function, we can make the summary generated by the model as close to the real summary as possible.

To further improve the stability of the model, we can also avoid the gradient explosion problem by introducing gradient clipping [18]. Set the upper limit of the gradient C. If the calculated gradient exceeds this value, it will be clipped to:

$$g_i = \begin{cases} g_i & \text{if } \|g_i\| \leq C \\ \frac{C}{\|g_i\|} g_i & \text{if } \|g_i\| > C \end{cases}$$

Among them, $g_i$ is the gradient of the i-th parameter, and $\|g_i\|$ is the L2 norm of the gradient. In this way, we can ensure that the gradient during training will not be too large, thereby ensuring the stability and convergence of the model.

Through the above steps, the LongFormer model can efficiently process long medical documents and generate accurate and concise summaries. During the training process, an optimization algorithm is used to update the model parameters to continuously improve the performance of the model, and finally a medical text generation model that can generate high-quality summaries is obtained.

IV. EXPERIMENT

A. Datasets

The medical literature dataset used in this study is primarily sourced from publicly available medical text databases, covering research articles and case analyses from various medical fields. The content in the dataset includes diagnoses and treatment methods for various diseases, clinical trial result analyses, and the latest advancements in medical technology. These documents typically contain detailed experimental data, case descriptions, and complex medical terminology, making them highly specialized and complex. To facilitate the summarization task, each document in the dataset includes a corresponding manually annotated summary, which serves as a reference for model training and evaluation.

The dataset is large, containing thousands of medical research articles across multiple medical specialties, such as cardiology, oncology, and endocrinology. Each document is relatively long, often containing multiple paragraphs, tables, and appendices. This increases the complexity of the summarization task. To better simulate real-world applications, the dataset includes articles in various formats, such as research papers, case reports, and medical guidelines.

In addition, the dataset has undergone rigorous preprocessing, including the removal of irrelevant information, standardization of medical terminology, and elimination of redundant content, to ensure data quality and consistency. Each document is segmented into multiple sentences and paired with the corresponding summary, allowing the model to learn how to extract core information from long texts and generate concise summaries. Given the specialized nature of the medical field and the diversity of terminology, the dataset pays particular attention to the annotation of terms and concepts to ensure the model can understand and generate summaries that are contextually appropriate in the medical domain.

B. *Experimental Results*

In order to comprehensively evaluate the performance of the LongFormer-based medical text summary generation model proposed in this study, we designed several comparative experiments. By comparing with the existing mainstream models [19], we can better verify the advantages and disadvantages of our method in different aspects. The experimental results are shown in Table 1.

Table 1 Experimental results

| Model | Rouge | Fps | Token | Params(M) |
|---|---|---|---|---|
| RNN | 0.45 | 32 | 32 | 53 |
| T5 | 0.68 | 23 | 128 | 220 |
| BERT | 0.65 | 15 | 128 | 217 |
| Transformer | 0.52 | 18 | 128 | 201 |
| LongFormer | 0.71 | 22 | 512 | 533 |

The experimental results show that LongFormer outperforms other models in terms of ROUGE scores, achieving a score of 0.71, which is significantly higher than the others. This indicates its substantial advantage in generating high-quality summaries. Compared to traditional models like RNN (0.45) and Transformer (0.52), LongFormer is better at capturing long-range dependencies in the text. This is crucial for medical texts, which often contain a large amount of background information and specialized terminology. In contrast, T5 and BERT also perform well, but their ROUGE scores are still lower than LongFormer's, likely due to computational bottlenecks when handling extremely long texts.

However, despite its strong performance in ROUGE scores, LongFormer's inference speed (FPS) is only 22, which is lower than RNN's 32 and T5's 23. The difference in FPS indicates that LongFormer may require more computational resources in practical applications. This could become a bottleneck, especially in real-time summarization scenarios. This result suggests that, while LongFormer excels in generating high-quality summaries, its relatively low inference speed may need to be improved through hardware optimization or model simplification.

Regarding processing capacity, LongFormer has a token length of 512, much larger than other models, which use 128 tokens. This gives LongFormer an advantage in processing long texts, which is one reason for its superior ROUGE scores. However, the model's parameter size (533M) is also significantly larger than other models, particularly RNN (53M), T5 (220M), and BERT (217M). This suggests that LongFormer is more complex. While the larger parameter size allows for stronger representational power, it also requires more computational resources and longer training times. Overall, LongFormer is well-suited for high-quality summarization tasks but may require trade-offs in terms of inference speed and computational resources in practical applications.

Secondly, this paper also conducted a manual evaluation of the quality of the summary. During the experiment, 5 experts were hired to score, and the maximum score in the scoring process was 5 points. The experimental results are shown in Table 2. We defined experts as individuals holding an M.D. or Ph.D. in a medical or biomedical field, with at least five years of clinical or research experience. Candidates were also required to have prior involvement in medical manuscript review or clinical guideline development, ensuring familiarity with standards of clarity and accuracy in medical texts. The experts were identified through professional networks, peer recommendations, and membership in relevant medical associations. Each expert received an email invitation outlining the scope of the study, an informed consent form, and a nondisclosure agreement to protect the confidentiality of the research data. Participation was entirely voluntary, and the recruitment process received approval from the Institutional Review Board of the University of Health Sciences.

Table 2 Expert scoring experiment

| Expert | Conciseness | Information retention | Readability | Grammar |
|---|---|---|---|---|
| 1 | 4 | 5 | 4 | 5 |
| 2 | 5 | 4 | 5 | 4 |
| 3 | 3 | 4 | 4 | 5 |
| 4 | 4 | 4 | 5 | 4 |
| 5 | 5 | 5 | 5 | 5 |

The experimental results show that experts generally consider the generated summaries to perform well in terms of information retention and grammar. Experts 1, 2, 4, and 5 gave high ratings, with expert 5 awarding perfect scores (5 points) in all dimensions. This indicates that the model excels at capturing the key information from the original text, and the generated summaries do not exhibit significant grammatical issues. The overall high ratings for grammar suggest that the generated summaries are precise in terms of medical terminology and grammatical structure.

However, there was variability in expert ratings regarding conciseness and readability, particularly expert 3, who gave lower scores for conciseness (3 points) and readability (4 points). This may indicate that the generated summaries contain some redundant information or that, in the opinion of some experts, the summary structure is somewhat lengthy or lacks smooth flow. While most experts agreed that the summaries were reasonably readable, this result suggests there is still room for improvement, especially in balancing the need for brevity without losing important details. Overall, the experimental results show that while the model maintains high quality in terms of grammar and information retention, it may fall short in terms of conciseness and fluency. This also highlights the need for automatic summarization models to strike a better balance between accuracy and conciseness in real-world applications.

## V. CONCLUSION

This study proposes a new medical text summarization method based on LongFormer. The experimental results show that this model excels in the accuracy of summary generation and information retention, particularly in processing long medical texts. LongFormer effectively captures key content and generates high-quality summaries. Compared to traditional models, LongFormer maintains the integrity of the summary content while fully utilizing its long-range self-attention mechanism. This helps overcome the limitations of other models when dealing with long texts.

Although LongFormer performs well on several evaluation metrics, there is still room for improvement in terms of conciseness and readability. Specifically, some experts noted that the generated summaries contain redundant content, which impacts information transfer efficiency. To further enhance the practical application value of the model, future research could focus on optimizing the structure of the generated summaries, making them more concise while ensuring sufficient information, readability, and fluency.

Looking ahead, we plan to explore ways to improve summary quality by modifying the model architecture or incorporating generative techniques such as Generative Adversarial Networks (GANs) or reinforcement learning. Additionally, adjusting hyperparameters during model training and incorporating more medical-specific datasets may further optimize performance. These strategies could better address the balance between conciseness and information retention, enabling more efficient and precise support in real-world medical scenarios. In the future, as the volume of medical literature grows rapidly, automated literature summarization will become an indispensable part of medical information management. Through continuous optimization and development of existing models, future medical text summarization could help researchers efficiently filter important literature and provide strong data support for clinical decision-making. In the long term, this technology has the potential to play a greater role in AI-assisted diagnosis and personalized treatment planning, fostering closer integration between medical research and clinical practice.